\documentclass{article} 
\usepackage{iclr2021_conference,times}
\usepackage{graphicx, xcolor}

\usepackage{amsmath,amsfonts,bm}

\def\eqref#1{equation~\ref{#1}}

\def\1{\bm{1}}

\def\vzero{{\bm{0}}}

\def\vtheta{{\bm{\theta}}}

\def\vv{{\bm{v}}}
\def\vw{{\bm{w}}}
\def\vx{{\bm{x}}}

\def\mI{{\bm{I}}}

\DeclareMathAlphabet{\mathsfit}{\encodingdefault}{\sfdefault}{m}{sl}
\SetMathAlphabet{\mathsfit}{bold}{\encodingdefault}{\sfdefault}{bx}{n}

\usepackage{hyperref}
\usepackage{url}

\title{A neural anisotropic view of underspecification in deep learning}

\iclrfinalcopy
\author{Guillermo Ortiz-Jimenez\\
Ecole Polytechnique F\'ed\'erale de Lausanne\\
Lausanne, Switzerland\\
\texttt{guillermo.ortizjimenez@epfl.ch} \\
\And
Itamar Franco Salazar Reque\thanks{Work done during an internship at Ecole Polytechnique F\'ed\'erale de Lausanne}\\
Universidad Nacional de Ingenier\'ia\\
Lima, Peru \\
\texttt{isalazarr@uni.pe} \\
\AND
Apostolos Modas\\
Ecole Polytechnique F\'ed\'erale de Lausanne\\
Lausanne, Switzerland\\
\texttt{apostolos.modas@epfl.ch} \\
\And
Seyed-Mohsen Moosavi-Dezfooli \\
ETH Z\"urich \\
Zurich, Switzerland\\
\texttt{seyed.moosavi@inf.ethz.ch} \\
\AND
Pascal Frossard\\
Ecole Polytechnique F\'ed\'erale de Lausanne\\
Lausanne, Switzerland\\
\texttt{pascal.frossard@epfl.ch}
}

\begin{document}

\maketitle

\begin{abstract}
The underspecification of most machine learning pipelines~\citep{damour2020underspecification} means that we cannot rely solely on validation performance to assess the robustness of deep learning systems to naturally occurring distribution shifts. Instead, making sure that a neural network can generalize across a large number of different situations requires to understand the specific way in which it solves a task. In this work, we propose to study this problem from a geometric perspective with the aim to understand two key characteristics of neural network solutions in underspecified settings: how is the geometry of the learned function related to the data representation? And, are deep networks always biased towards simpler solutions, as conjectured in recent literature?  We show that the way neural networks handle the underspecification of these problems is highly dependent on the data representation, affecting both the geometry and the complexity of the learned predictors. Our results highlight that understanding the architectural inductive bias in deep learning is fundamental to address the fairness, robustness, and generalization of these systems.
\end{abstract}

\section{Introduction}

Modern neural networks are brittle. Not only in the
adversarially robust sense, but also when tested under slightly different conditions than the ones used to validate their original performance \citep{geirhos2018TextureBias,hendrycksBenchmarkingNeuralNetwork2019,Taori2020}. In fact, despite great advances in the validation performance of neural networks, their lack of robustness to semantically meaningless distribution shifts remains, to this date, one of the greatest challenges in deep learning.

Arguably, the main reason for this slow progress lies in the fact that we do not know how to handle the underspecification of most machine learning pipelines, i.e., it is often possible to achieve the same performance in a held-out validation set using radically different predictors~\citep{damour2020underspecification}. For this reason, relying only on validation accuracy as sole design metric cannot inform of the robustness of a model to naturally occurring distribution shifts. Instead, making sure that a neural network can generalize under different testing conditions requires to understand the specific way in which it solves the task.

We propose to study the problem of underspecification from a geometric perspective with the aim to understand two key characteristics of the solutions found by neural networks in underspecified tasks. First, how is the geometric margin (i.e., distance to the decision boundary) of deep models related to the data representation? Understanding this property is fundamental to address the adversarial robustness of deep learning~\citep{proc_ieee}. And second, are deep networks always biased towards simpler solutions? This property has been conjectured to mediate in generalization both in- and out-of-distribution \citep{rahamanSpectralBiasNeural2018,nakkiran2019sgd,shah2020pitfalls}, but our results challenge the validity of this assumption.

Answering these questions, though, requires a deeper understanding of the inductive bias of deep neural networks, as it determines which features a neural network preferably picks to solve a task. Recently, it has been shown that, at least in the context of linearly separable tasks, the properties of the input gradients can partly characterize the inductive bias induced by the architecture~\citep{ortizjimenez2020neural}. In this work, we further show that these properties can also reveal new insights on the behaviour of neural networks in underspecified settings. Our results show that the way neural networks handle underspecification highly depends on the data representation, affecting both their geometric margin and the predictor complexity. We also show that under certain conditions, standard vision architectures can consistently solve underspecified tasks using complex and non-linear predictors, despite the existence of linear separators that can perfectly fit the training and test data. As far as we know, this is the first time that this type of behaviour is reported in the literature.

We see our work as a preliminary contribution to the understanding of how the architectural inductive bias deals with underspecification in deep learning. In particular, it highlights that addressing important challenges in fairness, robustness, and out-of-distribution generalization requires deeper insights into the inductive biases of these systems. Not only those sourcing from the data and learning algorithm, but also those driven by the preferences of the model, as well.

\section{Preliminaries}    

\begin{figure}
    \centering
    \includegraphics[width=0.8\textwidth]{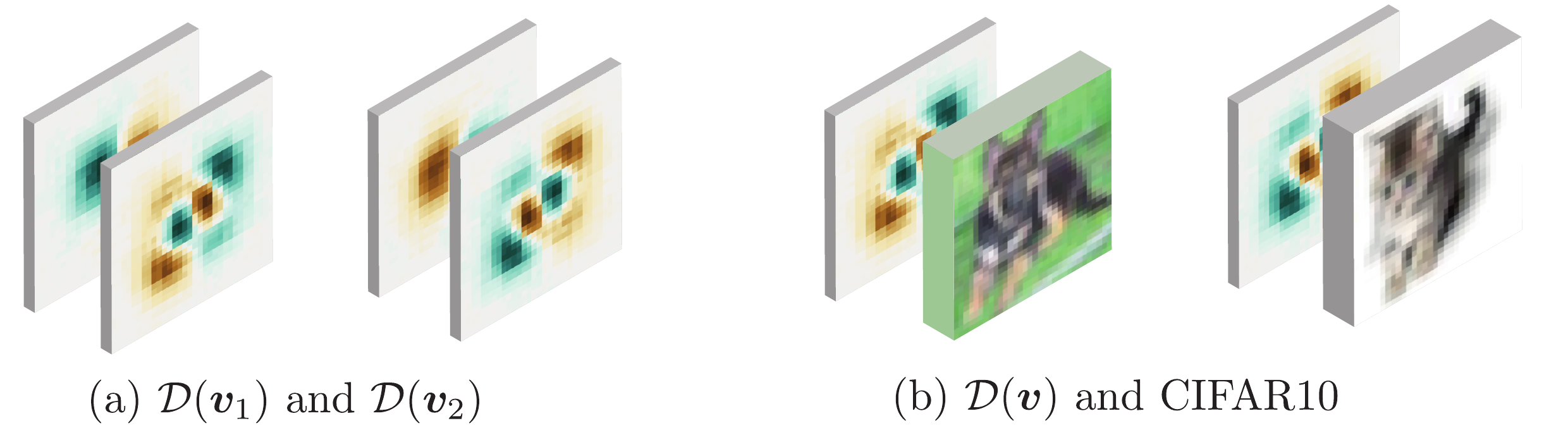}
    \caption{Diagram of the generation of the underspecified datasets. The samples from two datasets are concatenated along the channel dimension by stacking samples with the same label together.}
    \label{fig:datasets}
\end{figure}

Surprisingly, modern neural network architectures cannot generalize uniformly well on all linearly separable tasks \citep{ortizjimenez2020neural}.
Formally, one can construct the following families of linearly separable
data distributions $\mathcal{D}(\vv)$ with a single discriminative direction $\vv\in\mathbb{S}^{d-1}$ with margin $\epsilon>0$, such that for $(\vx,y)\sim\mathcal{D}(\vv)$,
\begin{equation}
    \vx=\epsilon \,y\vv+\vw\quad \text{with}\quad \vw\sim\mathcal{N}(\vzero, \sigma^2(\mI-\vv\vv^\top))\quad\text{and}\quad y\sim\mathcal{U}(\{-1,1\}).\label{eq:data_def}
\end{equation}
The performance of a network $f_\vtheta$, parameterized by a set of weights $\vtheta$, on $\mathcal{D}(\vv)$
is a monotonically increasing function of the alignment of $\vv$ with the input gradient covariance at initialization, i.e.,
\begin{equation}
    \alpha(\vv)=\vv^\top\left(\mathbb{E}_{\vtheta}\nabla_\vx f_\vtheta(\vx)\nabla^\top_\vx f_\vtheta(\vx)\right)\vv.
\end{equation}
This means that a neural network can only generalize well on linearly separable tasks whose discriminative direction is aligned with the largest eigenvectors of the gradient covariance. These eigenvectors are generally referred to as neural anisotropic directions (NADs) of a network, ordered in terms of their associated eigenvalues, i.e., lower/first NAD indices correspond to higher eigenvalues.

In this work, we use the concept of NADs to create different underspecified datasets consisting of the concatenation of data from two different distributions, $\mathcal{D}_1$ and $\mathcal{D}_2$, with the same number of classes. That is, we construct distributions of the form $\mathcal{D}=\mathcal{D}_1\oplus\mathcal{D}_2$, such that for every $(\vx,y)\sim\mathcal{D}$, we have $\vx=\vx_1\oplus\vx_2$ with $(\vx_1,y)\sim\mathcal{D}_1$ and $(\vx_2,y)\sim\mathcal{D}_2$. Here, $\oplus$ indicates the direct sum, or concatenation along the channel dimension, of two images (see Figure~\ref{fig:datasets}). To control the alignment with the inductive bias of the network, we always set $\mathcal{D}_1=\mathcal{D}(\vv)$ and choose $\vv$ to be one of the NADs of the network.

Note that this dataset construction is fully underspecified as the network can achieve optimal accuracy by using any of the channels, independently. Nevertheless, we expect that the trained model will use the information present in $\mathcal{D}(\vv)$ only when $\vv$ is aligned with the first NADs. In our experiments we test this hypothesis and use it to evaluate how the network behaves when trained on data that can be solved in more than one way.

\section{Neural anisotropies modulate input margin in deep networks}
\label{sec:3}

We first analyze the behaviour of a neural network when trained on a linearly separable, although underspecified, data distribution. Specifically, we generate data from $\mathcal{D}(\vv_1)\oplus\mathcal{D}(\vv_2)$ with $\epsilon_1=1$ and $\epsilon_2=0.5$, respectively. This dataset is underspecified as it can be perfectly separated using infinite linear separators in the span of $\vv_1$ and $\vv_2$. Nevertheless, even if all these solutions achieve the same validation performance, they can exhibit very different behaviours under distribution shifts.

\begin{figure}[t]
    \centering
    \includegraphics[width=0.9\textwidth]{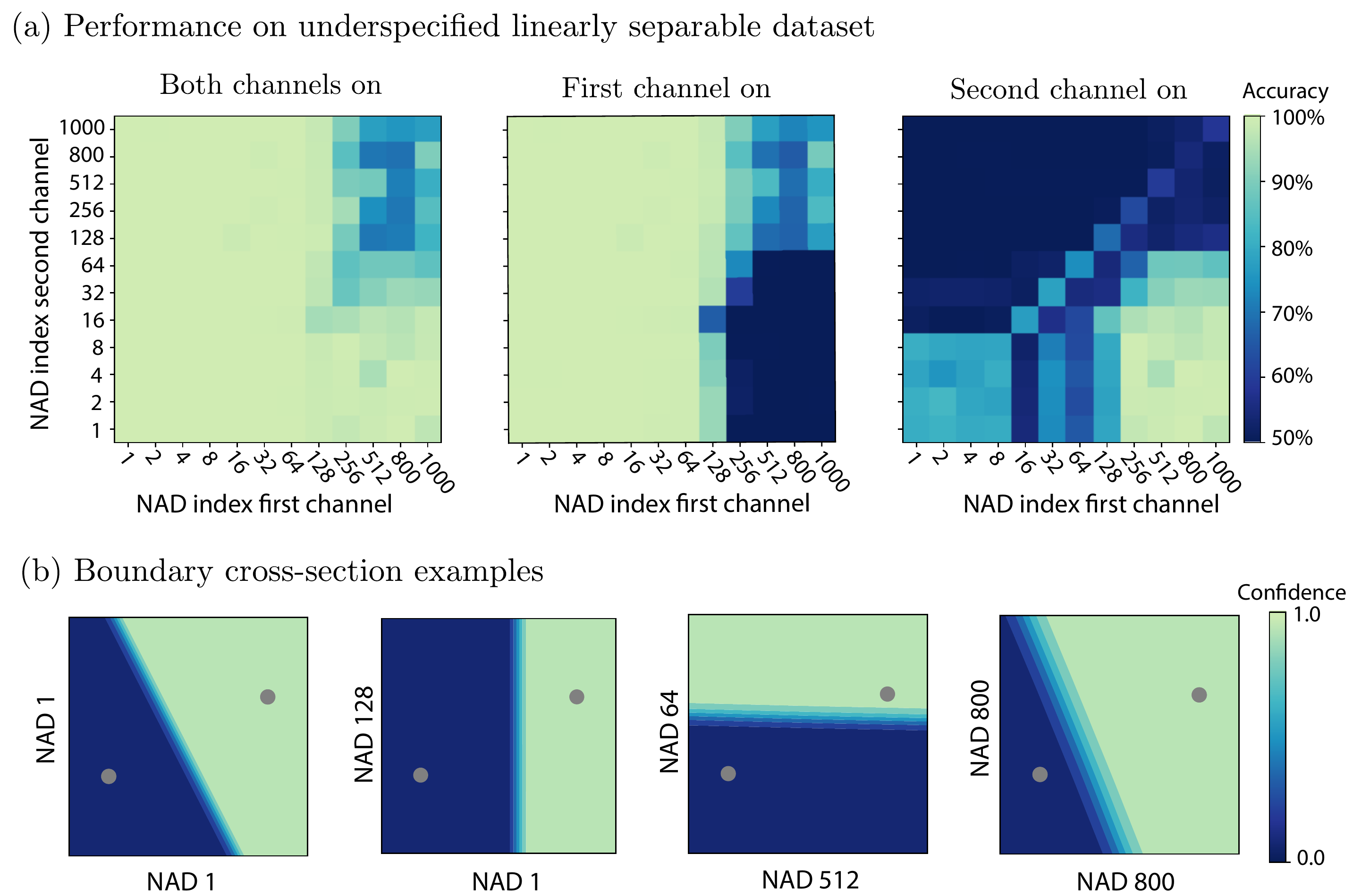}
    \caption{Analysis of the behaviour of a ResNet-18 on different versions of the underspecified linearly separable dataset with $\epsilon_1=1$ and $\epsilon_2=0.5$. Plots show average over three different random seeds. (a) In- and out-of-distribution generalization: (Left) Test accuracy on the original distribution with both channels on, (center) only the first channel on and second channel zeroed out, and (right) only the second channel on and first channel zeroed out. (b) Selected cross-sections of the decision boundary at a plane spanned by the first and second channel's discriminative directions.}
    \label{fig:linear_linear}
\end{figure}

Figure~\ref{fig:linear_linear}a summarizes the behaviour of a ResNet-18\footnote{Results with other networks can be seen in the Appendix.}~\citep{resnet} trained on different versions of the underspecified linearly separable dataset with varying degrees of alignment of $\vv_1$ and $\vv_2$ with the NAD basis. We observe that the overall performance of the network is poorer when both discriminative directions are aligned with the last NADs, in line with the results obtained when training on a single $\mathcal{D}(\vv)$~\citep{ortizjimenez2020neural}. On the other hand, we also see that, for many combinations of NAD indices, the network tends to use only the information of a single channel and completely neglects the information in the channel with the higher NAD index. This effect is also apparent when inspecting the tilt of the decision boundary in the plane spanned by $\vv_1\oplus\vzero$ and $\vzero\oplus\vv_2$, which ranges from being completely parallel to $\vv_1\oplus\vzero$, to being parallel to $\vzero\oplus\vv_2$.

The results of this experiment put forward the hardness of predicting the robustness properties of a neural network even in this very simple setup. Indeed, by changing the alignment of $\vv_1$ and $\vv_2$ with the NADs (amounting to a linear rotation of the input space) we see a whole array of different behaviours of the network in terms of margin: from behaving like a maximum input margin classifier when $\vv_1$ and $\vv_2$ are both aligned with the first NADs, to behaving almost like a decision stump in the NAD basis when either $\vv_1$ or $\vv_2$ are aligned with the last NADs~(see Figure~\ref{fig:linear_linear}b). Understanding how this phenomenon affects the robustness of deep networks and using it to boost their performance in standard settings seems, therefore, an important research direction for the future.

\section{Deep learning is not always biased towards simpler solutions}
\label{sec:4}
We now test a common conjecture about the inductive bias of neural networks, which postulates that deep networks tend to prefer to solve tasks with simpler solutions~\citep{rahamanSpectralBiasNeural2018}. In these terms the simplicity bias mostly refers to geometric simplicity of the decision boundary, e.g., it is suggested that neural networks will always prefer to separate a dataset linearly rather than non-linearly~\citep{nakkiran2019sgd}. This property is generally regarded as positive for generalization, however, it has also been shown to be possibly harmful for the robustness of these systems~\citep{shah2020pitfalls}. Contrary to this common assumption, however, we provide a counterexample in which neural networks do not abide by this supposed simplicity bias. 

\begin{figure}[t]
    \centering
    \includegraphics[width=\textwidth]{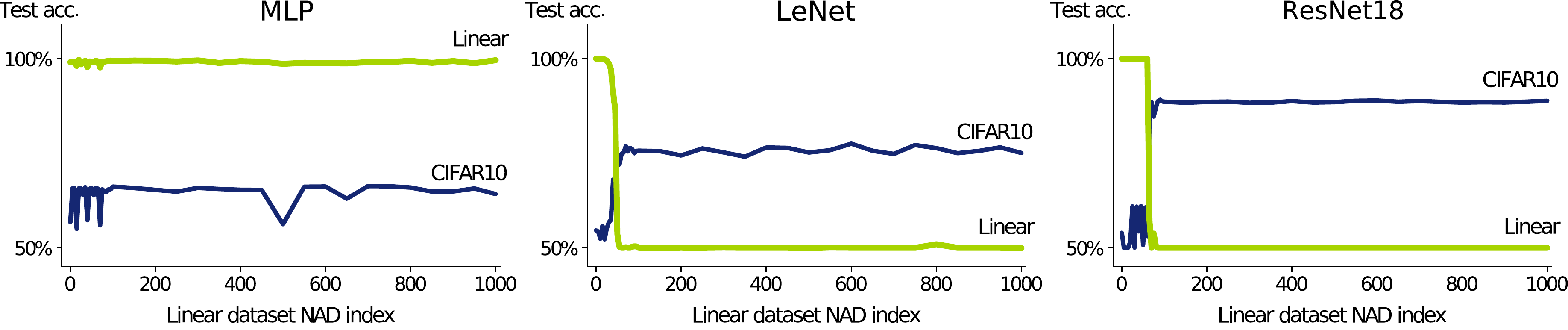}
    \caption{Out-of-distribution performance of an MLP, a LeNet~\citep{lecunGradientbasedLearningApplied1998}, and a ResNet-18~\citep{resnet} on different versions of the underspecified concatenation of binary CIFAR10 and a linearly separable dataset with a discriminative feature aligned with a certain NAD. Plots show average over three different random seeds.}
    \label{fig:linear_cifar}
\end{figure}

We construct this problem by expanding a binary version of CIFAR-10~\citep{krizhevskyLearningMultipleLayers2009} -- with one label for the first five CIFAR-10 classes, and another one for the rest -- with an extra channel distributed according to $\mathcal{D}(\vv)$. If the simplicity bias fully explained the inductive bias of deep learning, then any neural network would prefer to solve this dataset by constructing only a linear boundary to separate the information in the linearly separable channel. This seems to be the case for a multi-layer perceptron (MLP) (see Figure~\ref{fig:linear_cifar}) which always prefers to solve this task using the linearly rather than the non-linearly separable channel. However, this is not the case for the two CNNs in our experiments. Instead, there seems to exist a more nuanced behaviour of the simplicity bias for these models. Indeed, whenever $\vv$ is aligned with the first few NADs, these networks only fit the information in the $\mathcal{D}(\vv)$  channel -- as predicted by the simplicity bias. However, whenever $\vv$ is aligned with the higher NADs, the networks prefer to use only the CIFAR-10 data, despite this solution having worse test accuracy and, arguably, a much higher complexity than solving $\mathcal{D}(\vv)$.

On the other hand, even if one could always solve this dataset by fitting the information in all channels simultaneously, it seems that CNNs cannot do so. Instead, they seem to be able to either solve the task using the information in the linearly separable channel, or in the CIFAR-10 channels, exclusively. This again depends on the alignment of $\vv$ with the NADs. We, therefore, believe that understanding why this sudden change happens, and how the architectural inductive bias mediates the choice of features of the network, is another important avenue for future studies in the fairness and robustness of these systems.

\section{Conclusions}

We have shown that the way that neural networks handle underspecification in deep learning is heavily influenced by their architectural inductive bias. In particular, we have shown two simple examples in which the out-of-distribution generalization properties of a neural network are strongly determined by the input representation. In this sense, our results challenge certain common assumptions about the inductive bias of these systems, which tend to presume that the solutions found by neural networks are governed by basic geometric rules, e.g., simplicity bias. Instead, we have clearly shown that this mental picture, which might be true when dealing with fully connected architectures, is not fully representative of the type of biases present in modern architectures. 

We see our work as an important first step towards understanding how the architectural inductive bias of deep networks breaks underspecification in deep learning: It shows that the geometric inductive bias of deep networks is very nuanced, and that the coupling between input representation and model choice can significantly alter key properties of these systems. We believe that research in this directions will be, therefore, of paramount importance to address future challenges in fairness, robustness, and in and out of distribution generalization of deep networks.



\bibliography{iclr2021_conference}
\bibliographystyle{iclr2021_conference}

\appendix

\section{Experiment details}
\subsection{Concatenation of linearly separable distributions}
For the experiments of Sec.~\ref{sec:3} we use linearly separable distributions of the form of $\mathcal{D}(\vv)$ to create the synthetic datasets. The datasets consist of $10,000$ samples for training and $10,000$ samples for test, each sample shaped in the form of a $32\times32\times2$ image, with each channel distributed according to different NADs. We use $\sigma=1$ for all our experiments, and set $\epsilon_1=1$ and $\epsilon_2=0.5$, respectively.

We train the ResNet-18 for 30 epochs on batches of size 128, optimizing a standard cross-entropy loss function, using SGD with a triangular (cyclic) learning rate with maximum value 0.1, fixed momentum of $0.9$, and fixed weight decay of $10^{-5}$. We also train an MLP with 2 hidden layers with 100 and 20 neurons respectively, and a LeNet. These networks are trained similarly to ResNet-18 but for 50 epochs, with a maximum learning rate of 0.2 and no regularization. In every case, the networks achieve close-to-zero training loss.

\subsection{Linearly separable distribution appended to CIFAR-10 data}
For the experiments of Sec.~\ref{sec:4}, we first convert the 10-class CIFAR-10 dataset into a binary classification task. If $c\in\{0, \dots, 9\}$ denotes the CIFAR-10 classes, the labels $y\in\{-1,1\}$ of the binarized version are obtained as
\begin{align*}
    y = 
    \begin{cases} 
    -1, & \text{if} \quad c < 5 \\
    +1, & \text{if} \quad c \geq 5.
    \end{cases}
\end{align*}
The dimension of the CIFAR-10 RGB images is $32\times32\times3$, while the dataset consists of $50,000$ training and $10,000$ test images, equally distributed along the $10$ labels. Hence, $25,000$ training images are assigned the label $y=-1$ and $25,000$ the label $y=+1$. Similarly, for the test data, $5,000$ images are assigned the label $y=-1$ and $5,000$ the label $y=+1$.
Second, based on the setup of Sec.~\ref{sec:3}, we use linearly separable distributions of the form $\mathcal{D}(\vv)$ to create synthetic grayscale images of dimension $32\times32$. Again, these distributions are parameterized by a unit vector $\vv$ chosen from the NAD basis. Same as the CIFAR-10 binarization, the synthetic dataset consists of $50,000$ training and $10,000$ test images, where half of them are assigned the label $y=-1$ and the other half the label $y=+1$.
Finally, we concatenate the binarized CIFAR-10 dataset with the synthetically generated one, in order to create a 4-channel dataset.

We train on this 4-channel dataset an MLP ($60$ epochs) with 2 hidden layers of $200$ and $50$ neurons wide respectively, a LeNet ($40$ epochs), and a ResNet-18($40$ epochs) on batches of $128$. For every network, the discriminative feature size of the synthetic dataset is $\epsilon=1$, while the noise variance for the MLP and ResNet-18 is $\sigma=3.0$, and $\sigma=1.0$ for LeNet. The values of these parameters are set to obtain a clear phase transition between the two behaviours in our experiments. We optimize the standard cross-entropy loss, using SGD with a triangular (cyclic) learning rate with maximum value 0.15, fixed momentum of $0.9$, and fixed weight decay of $10^{-5}$. In every case, the networks achieve close-to-zero training loss.

\section{Additional results on underspecified linearly separable datasets}
\begin{figure}[t]
    \centering
    \includegraphics[width=0.9\textwidth]{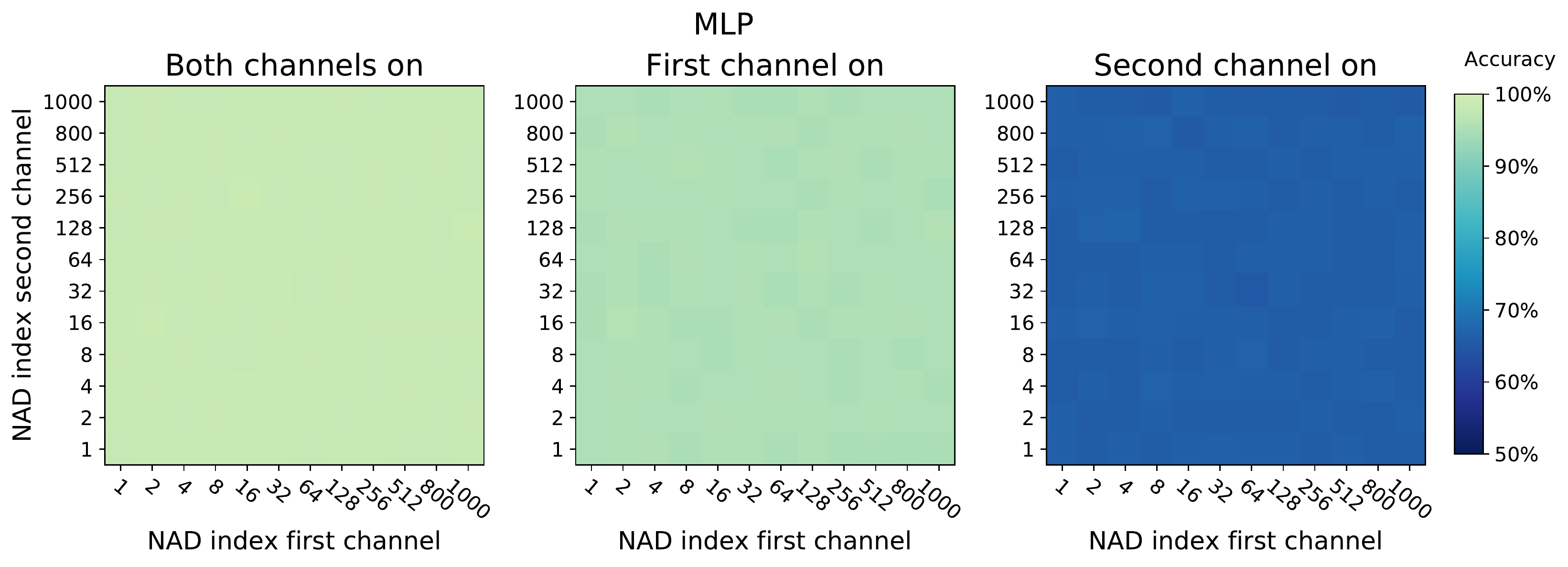}
    \caption{In and out-of-distribution generalization for MLP: (Left) Test accuracy on the original distribution with both channels on, (center) only the first channel on and second channel zeroed out, and (right)only the second channel on and first channel zeroed out.}
    \label{fig:underspecified_MLP}
\end{figure}

\begin{figure}[t]
    \centering
    \includegraphics[width=0.9\textwidth]{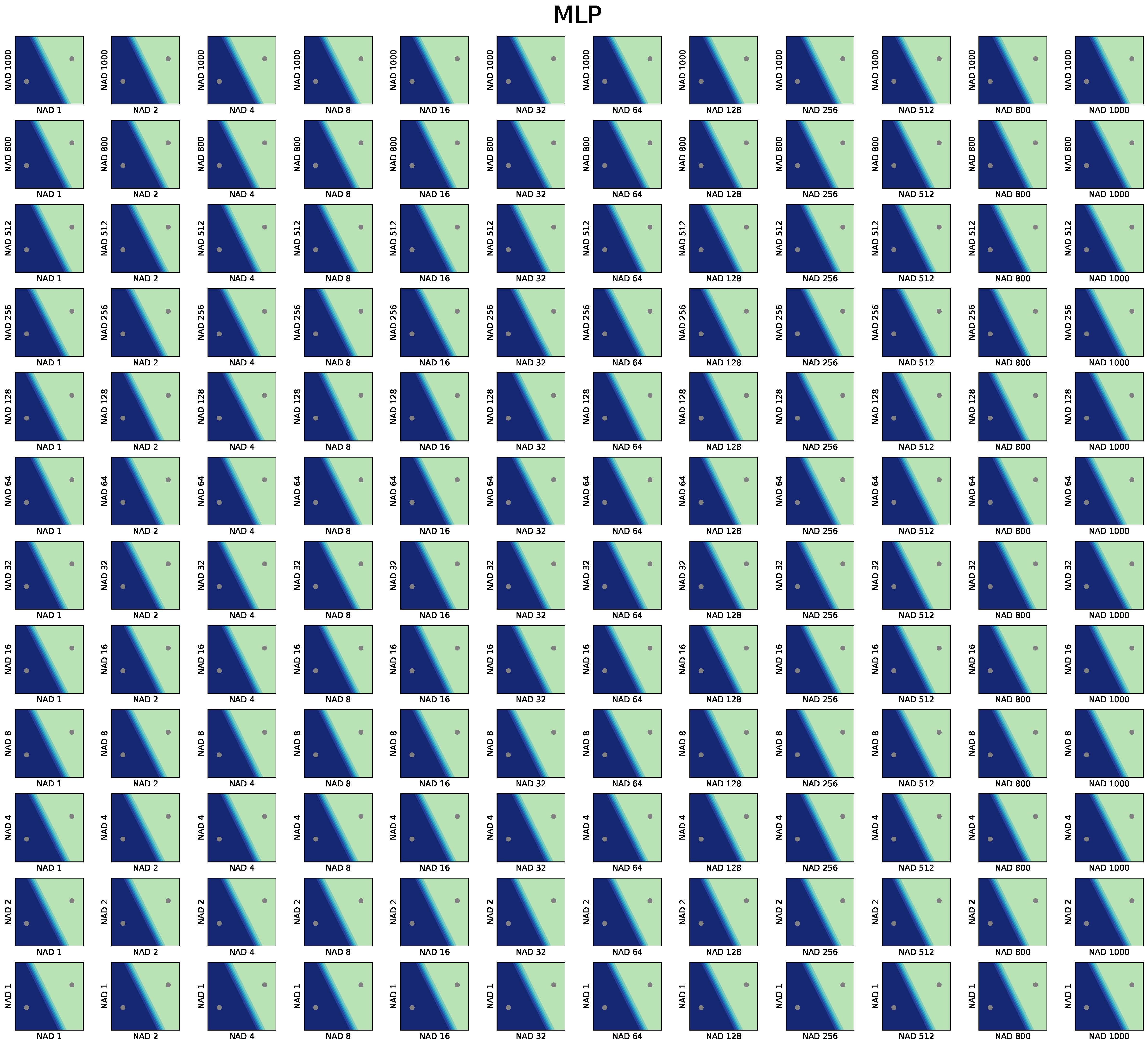}
    \caption{Cross-sections of the decision boundary at a plane spanned by the first and second channel’s discriminative directions for MLP}
    \label{fig:cross_MLP}
\end{figure}

We now show additional results on the underspecified linearly separable datasets with other networks. Figure~\ref{fig:underspecified_MLP} shows in and out-of-distribution generalization results for MLP. Note that MLP always achieves 100\% test accuracy, which drops uniformly when we only use one channel. Although the accuracy drop is smaller when we only use the first channel, probably due to the fact that the margin in this channel is bigger ($\epsilon=1.0$) and the network prefers to use this direction. This seems to be true when we inspect some cross-sections (see Figure~\ref{fig:cross_MLP}) of the decision boundary. Here, we see that the decision boundary in the plane spanned by the discriminative directions of the channels has a tilt mostly aligned with the first channel direction. We also see that these boundaries do not appear to change for any combination of NADs. All these behaviours contrast with LeNet and ResNet-18 networks.

\begin{figure}[t]
    \centering
    \includegraphics[width=0.9\textwidth]{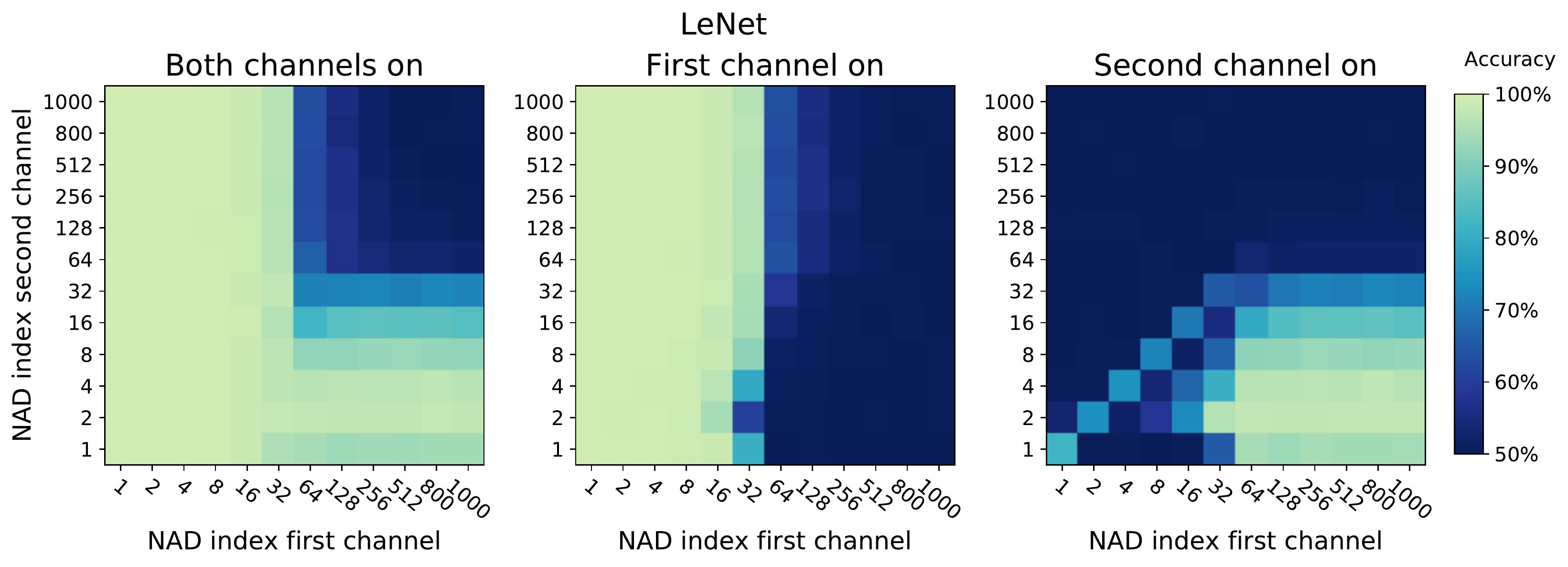}
    \caption{In and out-of-distribution generalization for LeNet: (Left) Test accuracy on the original distribution with both channels on, (center) only the first channel on and second channel zeroed out, and (right)only the second channel on and first channel zeroed out.}
    \label{fig:underspecified_LeNet}
\end{figure}

\begin{figure}[t]
    \centering
    \includegraphics[width=0.9\textwidth]{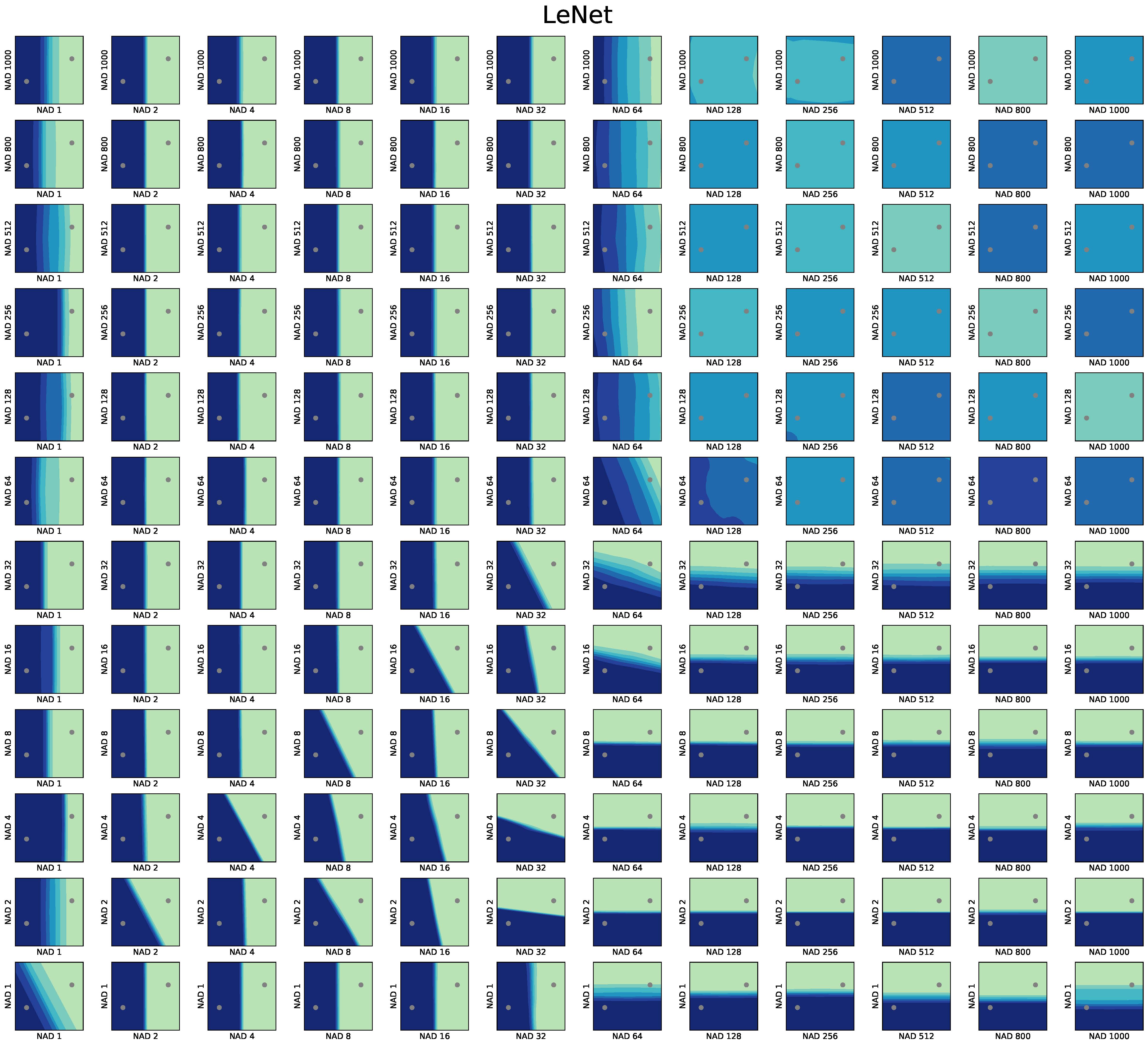}
    \caption{Cross-sections of the decision boundary at a plane spanned by the first and second channel’s discriminative directions for LeNet.}
    \label{fig:cross_LeNet}
\end{figure}

\begin{figure}[t]
    \centering
    \includegraphics[width=0.9\textwidth]{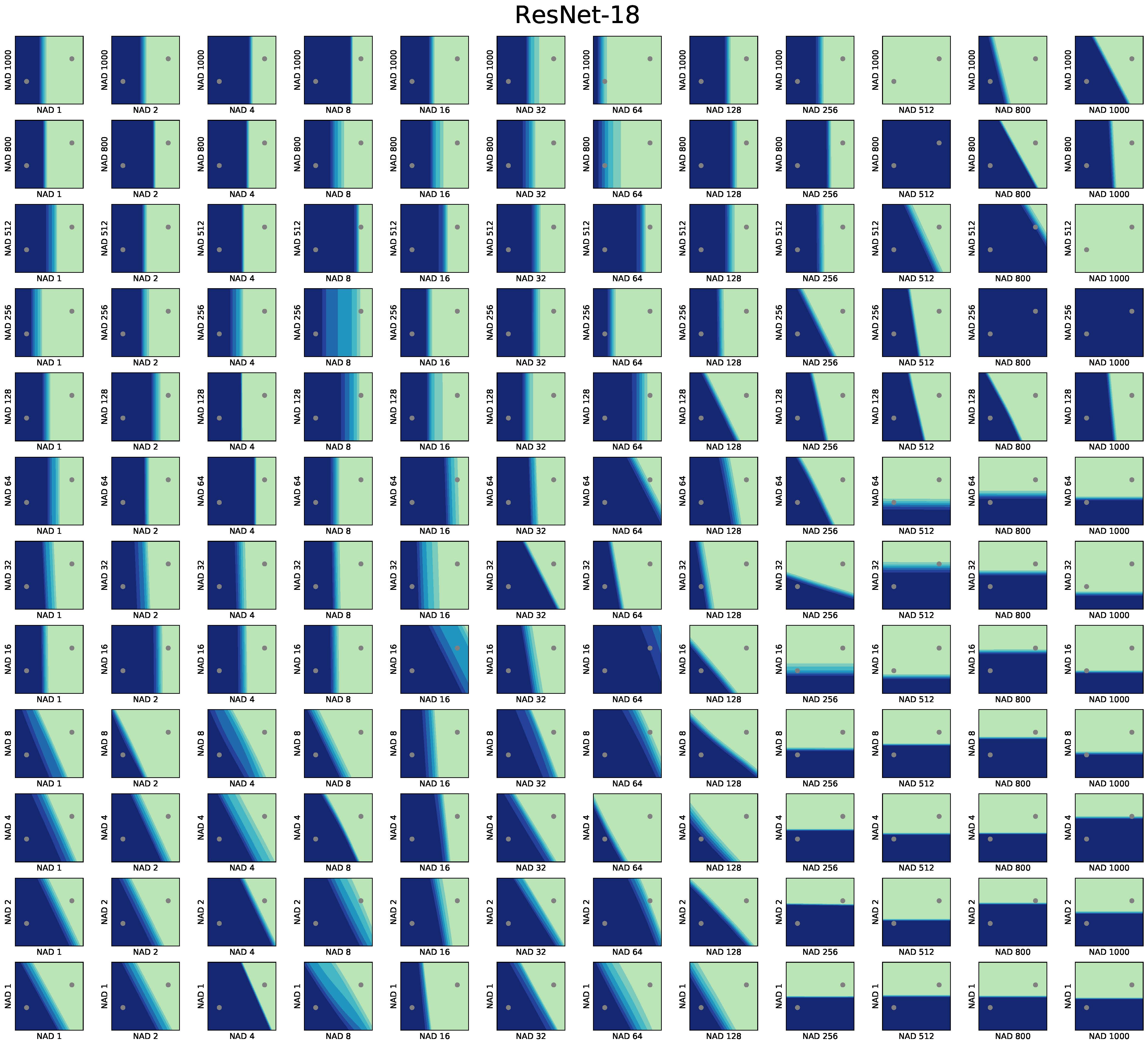}
    \caption{Cross-sections of the decision boundary at a plane spanned by the first and second channel’s discriminative directions for ResNet-18}
    \label{fig:cross_ResNet18}
\end{figure}

Figures~\ref{fig:underspecified_LeNet} and ~\ref{fig:cross_LeNet} show results for LeNet. In this case, as in ResNet-18, we observe how the overall performance of the network is poorer when the discriminative directions are aligned with the last NADs. As mentioned in Sec.~\ref{sec:3}, there seems to be a preference to use channels with discriminative directions aligned with first NADs even when their margins are smaller than those aligned with last NADs. Figure~\ref{fig:cross_LeNet} contains cross-sections for LeNet under different NAD pairs (see Figure~\ref{fig:cross_ResNet18} for ResNet-18). We can see how the decision boundary tilt changes from being completely parallel to the discriminative direction in the first channel, to being parallel to the discriminative direction used in the second channel and that this change is modulated by NADs.

\end{document}